\documentclass[10pt,twocolumn,letterpaper]{article}

\usepackage{iccv}
\usepackage{times}
\usepackage{epsfig}
\usepackage{graphicx}
\usepackage{amsmath}
\usepackage{amssymb}

\usepackage[breaklinks=true,bookmarks=false]{hyperref}

\iccvfinalcopy 

\def\iccvPaperID{****} 

\ificcvfinal\fi

\begin{document}

\title{Seek for Incantations: Towards Accurate Text-to-Image Diffusion Synthesis \\ through Prompt Engineering}

\author{Chang Yu\\
{\tt\small chang.yu@nlpr.ia.ac.cn}
\and
Junran Peng\\
{\tt\small jrpeng4ever@126.com}
\and
Xiangyu Zhu\\
{\tt\small xiangyu.zhu@nlpr.ia.ac.cn}
\and
Zhaoxiang Zhang\\
{\tt\small zhaoxiang.zhang@ia.ac.cn }
\and
Qi Tian\\
{\tt\small tian.qi1@huawei.com}
\and
Zhen Lei\\
{\tt\small zlei@nlpr.ia.ac.cn }
}

\maketitle
\ificcvfinal\thispagestyle{empty}\fi

\begin{abstract}
The text-to-image synthesis by diffusion models has recently shown remarkable performance in generating high-quality images. Although performs well for simple texts, the models may get confused when faced with complex texts that contain multiple objects or spatial relationships. To get the desired images, a feasible way is to manually adjust the textual descriptions, i.e., narrating the texts or adding some words, which is labor-consuming. In this paper, we propose a framework to learn the proper textual descriptions for diffusion models through prompt learning. By utilizing the quality guidance and the semantic guidance derived from the pre-trained diffusion model, our method can effectively learn the prompts to improve the matches between the input text and the generated images. Extensive experiments and analyses have validated the effectiveness of the proposed method.
\end{abstract}

\section{Introduction}
\begin{figure}[htbp]
    \centering
    \includegraphics[width=0.9\linewidth]{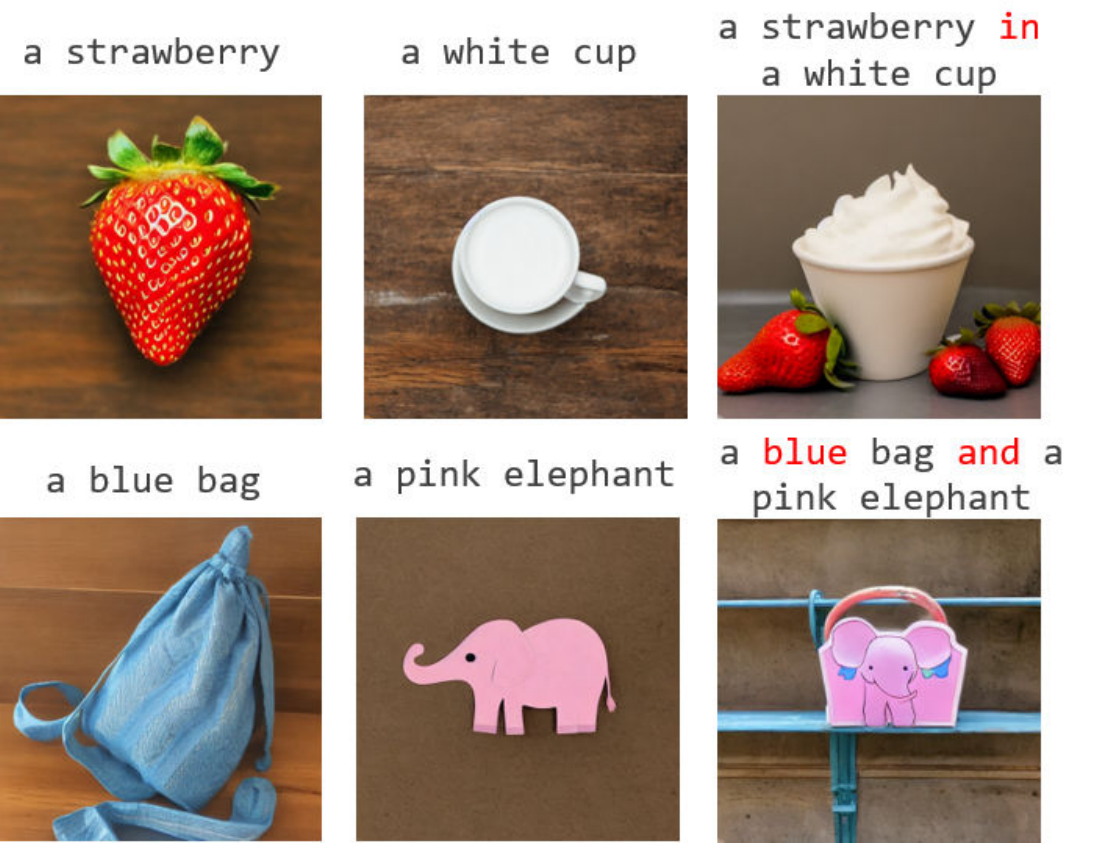}
    \caption{The text-to-image generation results of LDMs~\cite{rombach2022stable} with the same noise and the same random seed. It can be seen that the model performs well for short textual descriptions but degrades when the text becomes complex.}
    \label{fig:intro-fail}
\end{figure}

Recently, diffusion models such as DALLE-2~\cite{ramesh2022DALLE}, LDMs~\cite{rombach2022stable}, Imagen~\cite{saharia2022imagen}, etc., have shown remarkable results in generating high-fidelity images~\cite{ruiz2022dreambooth,liu2022compositional,hertz2022prompt,kawar2022imagic}. Given a short textual description, these models are able to generate the corresponding high-fidelity images based on the language model for text conditioning. However, when the texts are complex, i.e., containing multiple objects or with multiple attributes, the models tend to mix up their attributes and spatial relationships, or even miss some of the objects, leading to inaccurate generation results, as shown in Figure~\ref{fig:intro-fail}. 

\begin{figure}[htbp]
    \centering
    \includegraphics[width=1.0\linewidth]{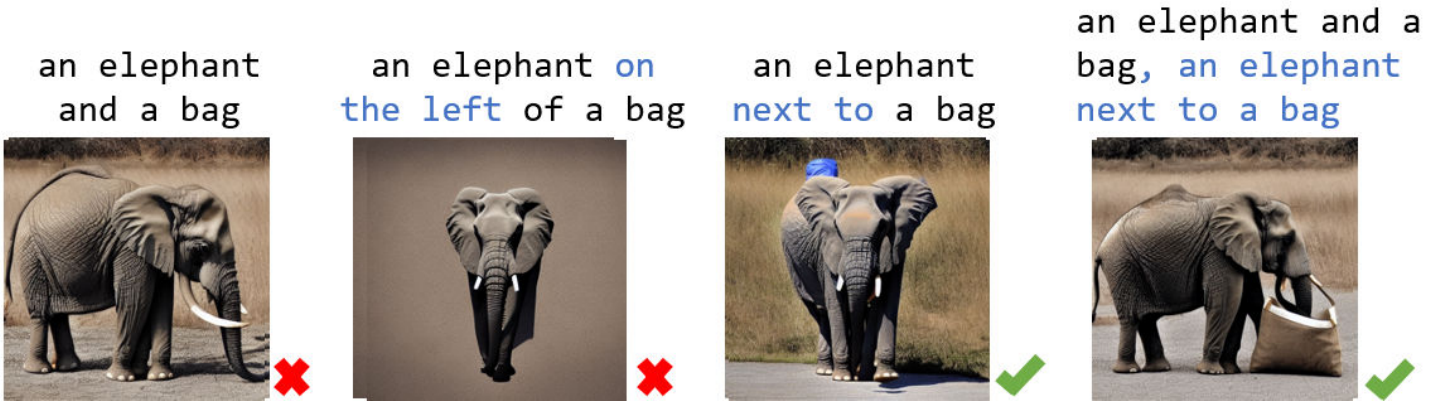}
    \caption{The generation results of LDMs~\cite{rombach2022stable} with different textual descriptions (same initial noise and same random seed). After carefully narrating the text or incorporating additional prompts, the model successfully synthesizes the images containing "an elephant and a bag".}
    \label{fig:intro-prompt}
\end{figure}

To tackle this issue, some researchers try to decompose the generation process into multi-steps. Liu et al.~\cite{liu2022compositional} propose to split the textual descriptions into a set of components, where each component is sent to the diffusion model independently and the estimated text-conditioned noises are summed at each sample step to generate the final outputs. The components of the method need to be defined in advance and given their positive or negative relationships. Feng et al.~\cite{feng2022training}   incorporate a language parser to decompose the text into structured representations and re-send the noun phrases in the text encoder for more accurate textual descriptions. The separately estimated text embeddings are combined at the cross-attention layers. These methods rely on the pre-decomposition of the input texts and the synthesis process has to be repeated for each component. Another feasible way to improve the generation results is to design the prompt of the text. When concatenating the input texts with special phrases or carefully narrating the texts, as shown in Figure~\ref{fig:intro-prompt}, the text-to-image matching will be improved. 
Based on this observation, many people try to explore the "Incantations" that enable effective synthesis~\cite{100prompts}. Although have achieved satisfying results, the selection of the prompts requires extensive manual adjustments and can only be applied to specific scenarios.

In this paper, we propose a 2-stage framework to improve the text-to-image diffusion synthesis through prompt engineering. The core of our method is to find a proper optimization direction to guide the learning of the prompt for each textual input so that the synthesized image is consistent with the text. The guidance of the prompt is derived from the trained diffusion models. Specifically, the random-sampled noise and the text are first sent to the diffusion model with a coarse step and a fine step separately to generate the coarse image and the fine image. Note that the diffusion model with more sampling steps usually performs better than it with fewer sampling steps, the direction from the coarse image to the fine image will provide quality guidance for prompt learning. Besides, taking into account the influence of narrating the text, part of the text is masked according to the similarity between the text and the image to enhance the semantic guidance for the learnable prompts. Combined with other consistency and sparsity constraints, our method effectively learns the proper prompts for each input text, which provides insight into the potential of pre-trained diffusion models to generate better text-conditioned images.

The main contributions are summarized as follows:
\begin{itemize}
\item This paper proposes a framework to improve text-to-image diffusion synthesis through prompt engineering. Without training the diffusion models, the proposed method utilizes the guidance derived from the diffusion model to seek for the prompts that make the text and the generated image match better.

\item The quality guidance derived from the trained diffusion models is proposed for prompt learning, which utilizes the direction from the coarsely-sampled image to the finely-sampled image to guide the optimization direction of the text with the prompt.

\item We propose to use the text-image similarity as semantic guidance to narrate the current text with prompts. By masking the words that get less attention, the semantic guidance will enhance the representation of those words through prompt learning. After training the prompts for a few iterations, the accuracy of the text-to-image synthesis is improved. Extensive experiments and analyses are conducted to validate the effectiveness of the proposed method.

\end{itemize}

\section{Related Work}
\subsection{Diffusion models}
Recently, models such as DALLE-2~\cite{ramesh2022DALLE}, Imagen~\cite{saharia2022imagen}, and LDMs~\cite{rombach2022stable} have attracted a lot of attention due to their amazing performance in generating creative images. Given an image sampled from its distribution, the image can be transferred to a Gaussian noise by adding Gaussian noise to the image gradually. The diffusion models execute the reverse process, which use a network to predict the added noise at each step to generate realistic images. Benefiting from the large-scale training datasets, the diffusion models have shown remarkable results in various tasks~\cite{yang2022diffusion}, including detection~\cite{chen2022diffusiondet}, segmentation~\cite{baranchuk2021label,graikos2022diffusion}, 3D reconstruction~\cite{lin2022magic3d,poole2022dreamfusion}, generation~\cite{rombach2022text,ruiz2022dreambooth,ho2022video}, etc. However, training an LDMs-like diffusion model from scratch requires a lot of computational resources. Thus, many works focus on incorporating pre-trained diffusion models to improve the downstream tasks. In this paper, we try to explore the potential of the pre-trained diffusion models in generating text-conditioned images.

\subsection{Text-to-Image Synthesis}
Due to the powerful models~\cite{radford2021CLIP,bao2021beit,wang2022beitv3} that bridge the gap between language and vision, the text-to-image synthesis field has achieved a certain breakthrough. Based on the diffusion models, the text-conditioned image synthesis has been applied to various scenarios, such as controllable image editing~\cite{hertz2022prompt,kawar2022imagic,wu2022unifying,nichol2021glide,avrahami2022blended}, inpainting~\cite{lugmayr2022repaint,saharia2022palette}, and image translation~\cite{meng2021sdedit,ruiz2022dreambooth}. Despite the satisfying performance on simple texts, current diffusion models tend to get confused when faced with complex texts. To tackle this issue, Liu et al.~\cite{liu2022compositional} propose Composable Diffusion to manually split the text into pieces and merge the estimated text-conditioned noises to generate the final outputs. However, Composable Diffusion is not suitable for texts with relationships. Thus, Feng et al.~\cite{feng2022structure} incorporate a language parser to decompose the texts and enhance the embedding of the noun words. Both of these two methods rely on the pre-processing of the input texts. Instead, we propose to use the learnable prompts to improve the diffusion models with complex textual input.

\label{overview}
\begin{figure*}[!htbp] 
    \centering
    \includegraphics[width=1.0\linewidth]{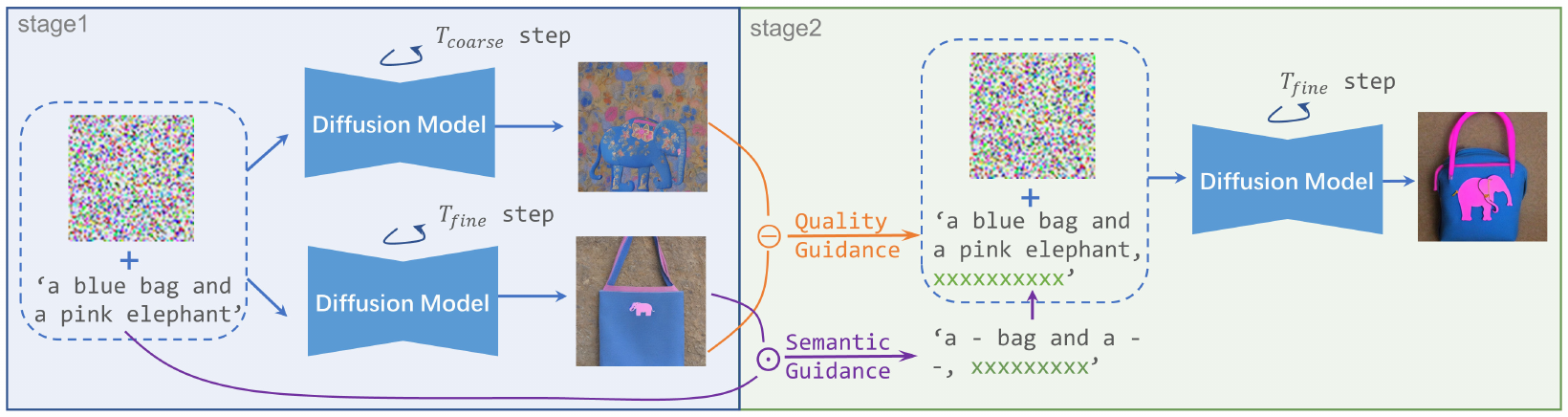}
    \caption{The overall framework of our method. The core of the method is to learn input-specific proper prompts for each textual input so that the generated images match well with the given texts. Firstly, the random-sampled noise $x_0$ and the original text are sent to the diffusion model with $T_{coarse}$ and $T_{fine}$ denoising steps separately. Afterward, the texts concatenated with the prompts are re-sent to the diffusion models with $x_0$ to generate the final outputs. During training, the difference between the coarsely-sampled images and the finely-sampled images is used as quality guidance to constrain the learning of prompts. Besides, the words that have lower similarity with the generated images are masked as semantic guidance to further enhance prompt learning. After training with consistency and sparsity constraints for a few iterations, the proposed method can effectively seek out the prompts to improve the text-to-image synthesis. }
    \label{fig:overview}
\end{figure*}

\subsection{Prompt Engineering}
The idea of prompt learning is first introduced to Natural Language Processing (NLP) field to enable the large models to be effectively adapted to downstream tasks~\cite{chen2022prompt,jiang2020can,shin2020autoprompt}. Inspired by the prompt engineering in NLP, CoOp~\cite{zhou2022learning} proposes to use similar technology in language-vision models. Instead of manually designing the prompts, CoOp models the prompts as a set of learnable parameters that can be concatenated with the original texts. By incorporating such automatic prompt engineering, CoOp successfully adapts to few-shot learning tasks. However, the prompts of CoOp are shared for all texts. To further improve the performance, CoCoOp~\cite{zhou2022conditional} proposes to use a meta-net to generate prompts according to the input texts. Note that the  manually-select prompts to improve text-conditioned diffusion models are labor-consuming. In this paper, we propose to use prompt engineering for better text-to-image synthesis, which provides insight into the potential of pre-trained diffusion models.



\section{Method}
Given a text description, a random-sampled noise, and a trained diffusion model,  our work aims to seek out the proper prompts to improve the accuracy of text-to-image synthesis. In the following, we will introduce the preliminary in Section~\ref{Preliminary}, our overall framework in Section~\ref{overview}, the details of the prompt engineering (including the quality guidance and the semantic guidance) in Section~\ref{promptengineer}, and the loss functions for prompt engineering in Section~\ref{loss}.

\subsection{Preliminary}
\label{Preliminary}
The diffusion model is based on the assumption that each image $x_0$ can be transferred to a Gaussian noise $x_T$ after gradually adding Gaussian noises $\epsilon$ to the image in $T$ times:
\begin{equation}\label{equ-pre-noise}
    q(x_t\vert x_{t-1}):=\mathcal{N}(\sqrt{1-\beta_t}x_{t-1},\beta_t\mathbf{I})),~t\in[1,T]
\end{equation}
where $\beta_t$ is the variance. As a result, the generation of images is regarded as a reverse sampling process, which uses the network $\epsilon_\theta(x_t,t)$ to estimate the added noise $\epsilon$ using the mean-squared loss:
\begin{equation}\label{equ-pre-denoise}
    E_{x,\epsilon\sim\mathcal{N}(0,1),t}\left[\Vert\epsilon-\epsilon_\theta(x_t,t)\Vert_2^2\right]
\end{equation}. In this paper, we denote the whole denoising process as $\Phi(x_T,T)$, the denoising process with condition $c$ as $\Phi(x_T,c,T)$.

\subsection{Overall Framework}
In this paper, we propose a 2-stage framework to improve the text-to-image diffusion synthesis through prompt engineering, which utilizes the quality guidance and the semantic guidance derived from the trained diffusion model to seek the proper prompts for each textual input. The overview of the framework is shown in Figure~\ref{fig:overview}.

In the first stage, the text is sent to a pre-trained text encoder $\xi_{txt}(\cdot)$ to get the word-specific textual embedding $e_{txt}\in \mathcal{R}^{n \times d}$, where $n$ is the length of the text and $d$ is the dimension of the embedding. Then, the random-sampled noise $x_T$ and the text embedding $e_{txt}$ are sent to the pre-trained diffusion model $\Phi$ with $T_{coarse}$ and $T_{fine}$ sampling steps separately to generate the synthesis images $x^{coarse}_0$ and $x^{fine}_0$, which are combined with the text embedding to provide quality guidance and semantic guidance for prompt learning. The first stage is formulated as:
\begin{equation}\label{equ-stage1}
\begin{aligned}
    &e_{txt} = \xi_{txt}(text),~~x_T \sim \mathcal{N}(0,\mathcal{I}) \\
    &x^{coarse}_0 = \Phi(x_T, e_{txt},T_{coarse}),\\
    &x^{fine}_0 = \Phi(x_T, e_{txt},T_{fine}).\\ 
\end{aligned}
\end{equation}

In the second stage, the input text is concatenated with the learnable prompt $e_{prompt}$ and then sent to the text encoder $\xi_{txt}(\cdot)$ to form the final text embedding $\hat{e}_{txt}$. Afterward, $\hat{e}_{txt}$ and the noise $x_T$ in the first stage are sent to the pre-trained diffusion model with $T_{fine}$ sampling steps to synthesize the final output image $\hat{x}_0$:
\begin{equation}\label{equ-stage2}
\begin{aligned}
    &\hat{e}_{txt} = \xi_{txt}([text,e_{prompt}]) \\
    &\hat{x}_0 = \Phi(x_T, \hat{e}_{txt},T_{fine}).\\ 
\end{aligned}
\end{equation}
Aftering training the prompt $e_{prompt}$ with the guidance derived from the diffusion model, the text concatenated with the prompt will generate satisfying results. The guidance for prompt engineering will be introduced below. 

\begin{figure}[!htbp] 
    \centering
    \includegraphics[width=1.0\linewidth]{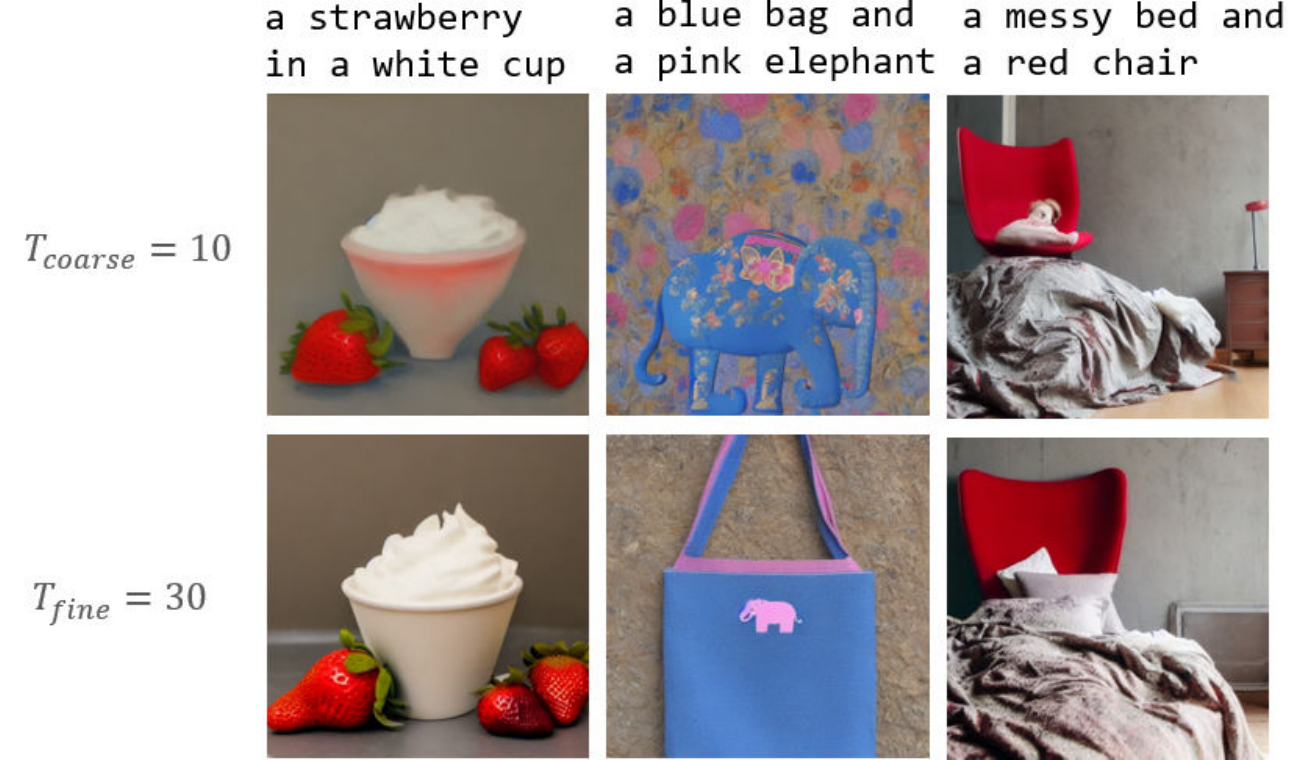}
    \caption{The influence of the sampling steps on the results of diffusion models. It can be seen that the results under more sampling steps ($T_{fine}$) are of better 'quality' than the ones under fewer sampling steps ($T_{coarse}$). The 'quality' includes how the text matches the image and whether the image contains distortion or artifacts.  }
    \label{fig:method-step}
\end{figure}

\begin{figure}[!htbp]
    \centering
    \includegraphics[width=0.7\linewidth]{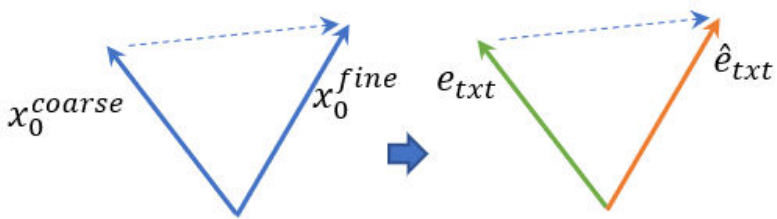}
    \caption{Illustration of the Quality Guidance. It incorporates the direction from the coarsely-sampled image to the finely-sampled image to guide the learning of the prompts.}
    \label{fig:method-quality}
\end{figure}

\subsection{Prompt Engineering}
\label{promptengineer}
The core of the proposed method is to learn the proper prompt for each textual input to make the synthesized image and the text match better. In this section, we will introduce how to derive the optimization direction from the trained diffusion models to guide the learning of prompts.

\noindent\textbf{Quality Guidance.} To learn the proper prompt for each text, we need to point out an optimization direction that will improve the text-to-image synthesis. For the denoising process of diffusion models, one common sense is that the results under more sampling steps are of better quality than the ones under fewer sampling steps. The results denoising from the same init noise under different sampling steps are shown in Figure~\ref{fig:method-step}. It can be seen that the 'quality' includes how the text matches the image and whether the image contains distortion or artifacts. For example, the finely-sampled image in the first column is clearer than the coarsely-sampled image, the finely-sampled image in the second column contains more accurate content and the finely-sampled image in the third column fits common knowledge better. Based on this observation, we propose to use the direction from the coarsely-sampled image to the finely-sampled image to guide the learning of the prompts:
\begin{equation}\label{equ-quality}
\begin{aligned}
    &e^{coarse}_{img} = \xi_{img}(x^{coarse}_0),~e^{fine}_{img} = \xi_{img}(x^{fine}_0),\\
    &\Delta{e_{img}} = \frac{e^{fine}_{img}}{\Vert e^{fine}_{img} \Vert_2} - \frac{e^{coarse}_{img}}{\Vert e^{coarse}_{img} \Vert_2}, \\
    &\Delta{e_{txt}} = \frac{\hat{e}_{txt}^g}{\Vert \hat{e}_{txt}^g \Vert_2} - \frac{e_{txt}^g}{\Vert e_{txt}^g \Vert_2},\\
    &\mathcal{L}_{qual} = \Delta{e_{img}}\cdot\Delta{e_{txt}},\\
\end{aligned}
\end{equation}
where $e^{coarse}_{img}$ and $e^{fine}_{img}$ are the image embedding extracted using the image encoder $\xi_{img}(\cdot)$, where $\hat{e}_{txt}^g\in\mathcal{R}^{1\times d}$ and $e_{txt}^g\in\mathcal{R}^{1\times d}$ are the global text embedding mapped from $\hat{e}_{txt}^g$ and $e_{txt}^g$ following~\cite{radford2021CLIP}. As shown in Figure~\ref{fig:method-quality}, $\mathcal{L}_{qual}$ constraints the consistency between the coarsely-sampled image to the finely-sampled and the text embedding to the text-with-prompt embedding, which provides quality guidance for the learning of prompts.

\begin{figure}[!htbp] 
    \centering
    \includegraphics[width=1.\linewidth]{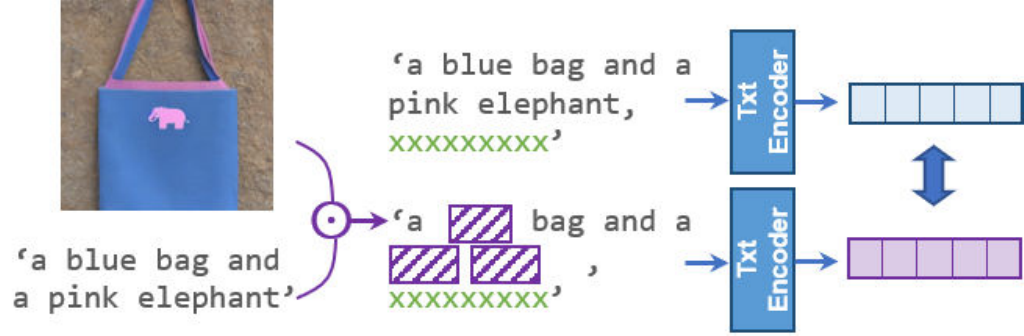}
    \caption{Illustration of the Semantic Guidance. The word-specific text-image similarities are used as the guidance to narrate the current text with prompts, which will enhance the learning of the contents that are with less attention before.}
    \label{fig:method-semantic}
\end{figure}

\noindent\textbf{Semantic Guidance.} The way how the text is described matters the generation results. As shown in Figure~\ref{fig:intro-prompt}, when we want a picture containing a bag and an elephant, 'an elephant and a bag' will lead to the neglect of the bag but 'an elephant next to a bag' or 'an elephant and a bag, an elephant and a bag' will generate the desired image. Note that the text encoder models the relationships between the words, we propose to use the text-image similarity as semantic guidance to narrate the current text with prompts, which will enhance the learning of the contents that are with less attention before. The detailed illustration for semantic guidance is shown in Figure~\ref{fig:method-semantic}. Given an output image $x_0$ and the original text, the word-specific similarity score $s\in\mathcal{R}^{n}$ is obtained. For the $R_i$ lower than the threshold $\gamma$, its corresponding word is replaced with '-' to get masked. Both the original text and the masked text are concatenated with the same prompt and sent to the text encoder $\xi_{txt}(\cdot)$. Then, their similarity are measured as:
\begin{equation}\label{equ-semantic}
\begin{aligned}
    &\hat{e}_{txt}^{mask}=\xi_{txt}(masked~text),\\
    &\mathcal{L}_{sem}=\hat{e}_{txt}\cdot\hat{e}_{txt}^{mask}. \\
\end{aligned}
\end{equation}
By constraining the semantic consistency between the original text and the masked text, semantic guidance $\mathcal{L}_{sem}$ enhances the learning of the neglected words, which will improve the results of text-to-image synthesis.

\begin{figure*}[!htbp] 
\centering
\includegraphics[width=1.0\linewidth]{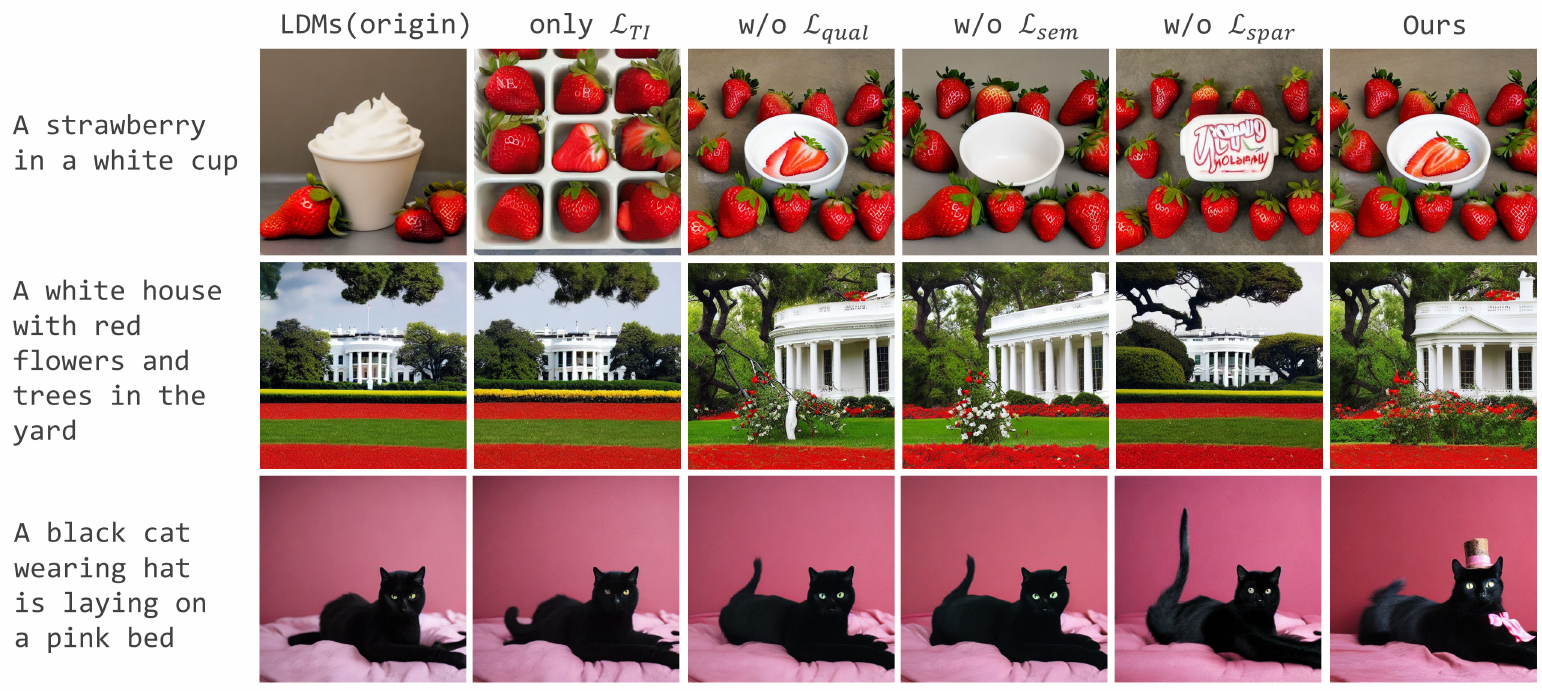}
\caption{The qualitative ablation study of the proposed method. It can be seen that each component contributes to generating satisfying images that match the given textual descriptions.}
\label{fig:abla-loss}
\end{figure*}


\subsection{Loss Functions}
\label{loss}
Despite the quality guidance and the semantic guidance, the remaining loss functions to train the prompt will be introduced in the following:

\noindent\textbf{Text-Text.} To avoid textual semantic drift during prompt learning, we constrain the semantic consistency between the original text and the text with prompts:
\begin{equation}\label{equ-loss-tt}
\mathcal{L}_{TT}=\frac{\hat{e}_{txt}^g}{\Vert \hat{e}_{txt}^g \Vert_2}\cdot\frac{e_{txt}^g}{\Vert e_{txt}^g \Vert_2}.\\
\end{equation}

\noindent\textbf{Text-Img.} Similarly, the semantic consistency between the generated image $x_0$ and the original text is used to void the semantic drift of the images:
\begin{equation}\label{equ-loss-ti}
\mathcal{L}_{TI}=\frac{\hat{e}_{txt}^g}{\Vert \hat{e}_{txt}^g \Vert_2}\cdot\frac{e_{img}}{\Vert e_{img} \Vert_2},\\
\end{equation}
where $e_{img}=\xi_{img}(x_0)$ is the image embedding encoded by the image encoder $\xi_{img}(\cdot)$.

\noindent\textbf{Sparsity.} Recent work has shown that the repeated prompt might affect the details of the background. To avoid learning the same repeated words as well as enable the prompt to learn effective content, we constrain the semantic sparsity of the learned prompt:
\begin{equation}\label{equ-loss-sparsity}
\mathcal{L}_{spar}=\sum_{i=n_o+1}^{n_o+n_p}\sum_{j=n_o+1, i\neq j}^{n_o+n_p}\left\vert \frac{\hat{e}_{txt}^i}{\Vert \hat{e}_{txt}^i \Vert_2}\cdot\frac{\hat{e}_{txt}^j}{\Vert \hat{e}_{txt}^j \Vert_2} \right\vert_1,\\
\end{equation}
where $\hat{e}_{txt}^i$ denotes the $i$th textual embedding that corresponds to the $i$th word of the text, $n_o$ is the length of the original text, and $n_p$ represents the length of the prompt.

The final loss functions for prompt engineering are combined as:
\begin{equation}\label{equ-loss-final}
\begin{aligned}
\mathcal{L}=&\lambda_{qual}\mathcal{L}_{qual}+\lambda_{sem}\mathcal{L}_{sem} + \lambda_{TT}\mathcal{L}_{TT} +\lambda_{TI}\mathcal{L}_{TI}\\
& +\lambda_{spar}\mathcal{L}_{spar},\\
\end{aligned}
\end{equation}
where $\lambda_{qual},\lambda_{sem},\lambda_{TT},\lambda_{TI}$, and $\lambda_{spar}$ are the hyper-parameters to balance different loss functions.


\section{Experiments}
\subsection{Implement Details}                                                       
The only learnable parameters in our framework are the prompts, which are specific to each textual input. We build our framework on the pre-trained LDMs~\cite{rombach2022stable} model, which incorporates the image encoder and text encoder from CLIP~\cite{radford2021CLIP} to extract embeddings. For the optimization, we use the Adam optimizer~\cite{kingma2014adam} with $10^{-2}$ learning rate to train the prompts, which are concatenated after the original texts. More training and evaluation details are provided in the supplementary material.

\begin{figure*}[!htbp] 
\centering
\includegraphics[width=1.\linewidth]{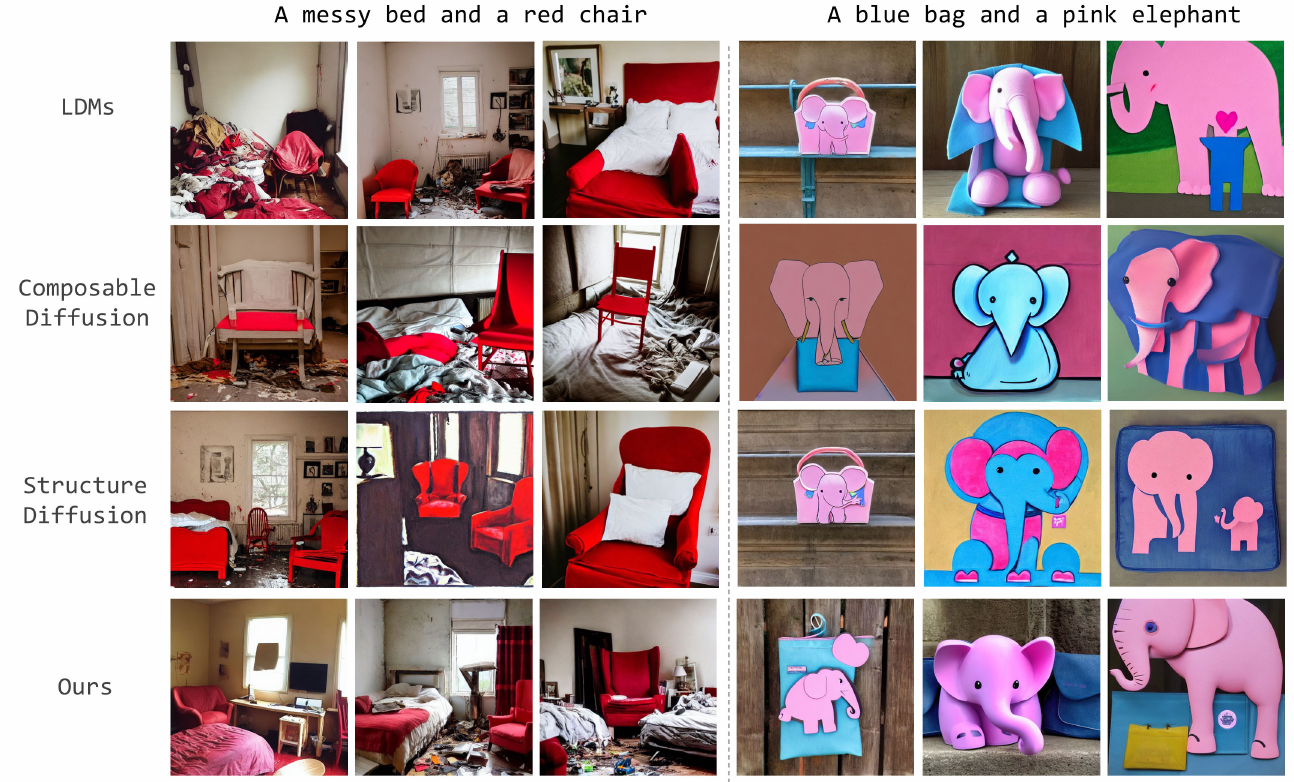}
\caption{The qualitative comparison of the images generated with the composable texts. Each column is generated with the same random-sampled noise with the same seed. It can be seen that our method generates more accurate images based on the given texts.}
\label{fig:compare-compose}
\end{figure*}
\subsection{Ablation Study}
The core of the paper is to find effective guidance to learn the prompt so that the texts and the generated images match better. From intuition, $\mathcal{L}_{TI}$ that constrains the consistency between the text and the image should be effective to improve the generation. However, note that the generation of diffusion is an iterative process, optimizing the generated image will result in the computational consumption for gradient backpropagation. In this paper, we propose a framework that freezes the diffusion model and use the prompt to explore the potential of the pre-trained diffusion model in text-to-image synthesis.

To validate the effectiveness of our method, we conduct the ablation study on the components of the framework, including the quality guidance $\mathcal{L}_{equal}$ (see Eqn.~\ref{equ-quality}), the semantic guidance $\mathcal{L}_{sem}$ (see Eqn.~\ref{equ-semantic}) and the sparsity constraint $\mathcal{L}_{spar}$ (see Eqn.~\ref{equ-loss-sparsity}). The results are shown in Figure~\ref{fig:abla-loss}. Comparing the first column and the second column, it can be seen that only constraining the consistency between texts and images will not obviously improve the results. The learning of the prompts needs more information. For the quality guidance $\mathcal{L}_{equal}$, it can be seen from the third column that the quality derived from the pre-trained models contributes to both the content-matching and the artifact-reducing. Specifically, without $\mathcal{L}_{equal}$, the strawberry inside the cup in the first row seems not realistic, the column of the house in the second row has artifacts when interacting with the flowers, and the image in the third row misses the hat. For the semantic guidance $\mathcal{L}_{sem}$, the results of the fourth column and the last column show that it effectively helps the network to focus on the contents that have been neglected before, leading to more accurate text-to-image synthesis. For the sparsity constraint $\mathcal{L}_{spar}$, it regularizes the space for searching the suitable 'incantations' for each textual input, which forces the network to learn the prompts effectively. The effect of $\mathcal{L}_{spar}$ is validated by comparing the last two columns of Figure~\ref{fig:abla-loss}. To sum up, it can be concluded that each component of the framework contributes to generating satisfying images, which explores the potential of the pre-trained diffusion models to synthesize the images that match the given textual descriptions better.                 
\begin{figure*}[!htbp]                      
\centering
\includegraphics[width=1.\linewidth]{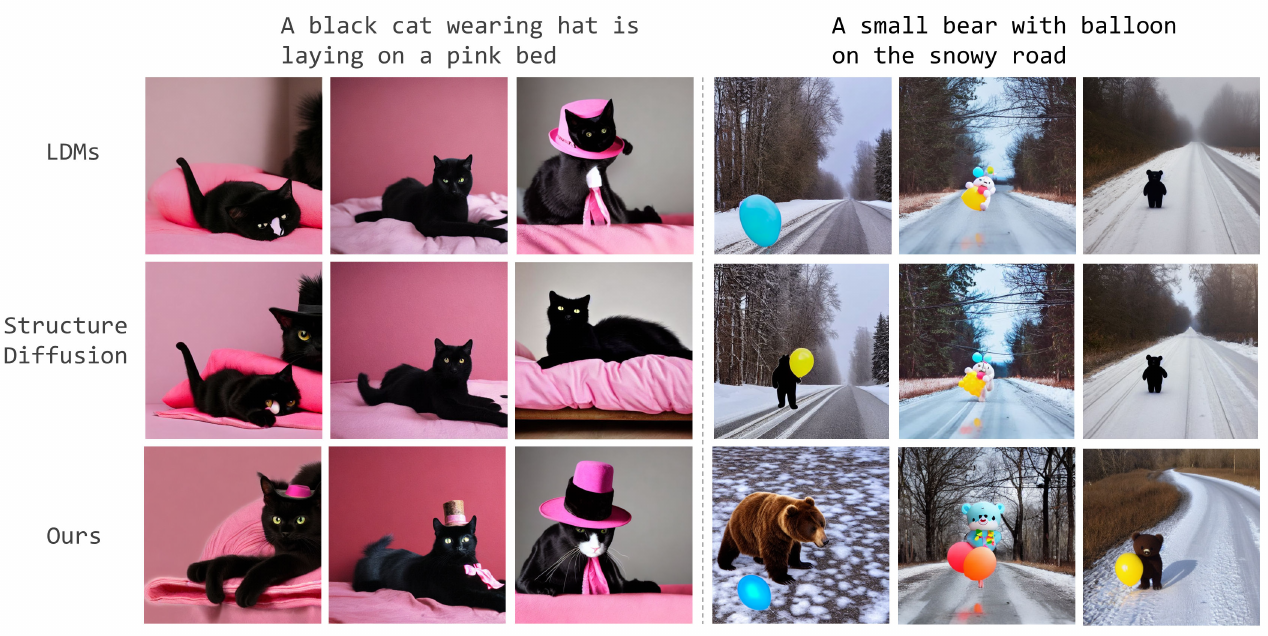}
\caption{The qualitative comparison of the images generated with the relational texts. Each column is generated with the same random-sampled noise with the same seed. It can be seen that our results match better with the given texts.}
\label{fig:compare-relation}
\end{figure*}

\subsection{Composable Text}
\label{composable}
Generally, text descriptions can be categorized into two types. One is the composable text, which focuses on the composition of its components, i.e., “A bag and an elephant”. The generated images are expected to include all of the mentioned objects, but indifferent to the relations between them. The other is the relational text, which not only cares about the objects but also the spatial or other relationships between them. The task of synthesizing images with complex text has not been well explored. Current methods~\cite{liu2022compositional,feng2022structure} mainly focus on the former one. To be specific, Composable Diffusion~\cite{liu2022compositional} proposes to manually split the text into pieces and merge the estimated text-conditioned noises to generate the final outputs. Structure Diffusion~\cite{feng2022structure} states that the unsatisfying results of the original diffusion models are mainly due to the limitation of the text encoder. Thus, Structure Diffusion proposes to incorporate a language parser to decompose the texts and enhances the embedding of the noun words. 

Different from these methods, we try to explore the potential of the diffusion models, which uses the guidance derived from the pre-trained diffusion model to learn the text-specific prompt for better text-to-image synthesis. The comparison results between the aforementioned methods and our method on composable texts are shown in Figure~\ref{fig:compare-compose}. It can be seen that when the origin diffusion model (LDMs~\cite{rombach2022stable}) gets defused when faced with complex texts. As for Composable Diffusion, the synthesized images tend to mix the two objects together, leading to inaccurate contents. This is because it roughly merges the estimated components regardless of their spatial locations. For Structure Diffusion, the improvement is limited as the inaccuracy from the text encoder is not the only reason for inaccurate results. By comparison, our method performs better, as it uses the guidance derived from the diffusion model to learn the effective input-specific prompts.

\begin{figure*}[!htbp] 
\centering
\includegraphics[width=1.0\linewidth]{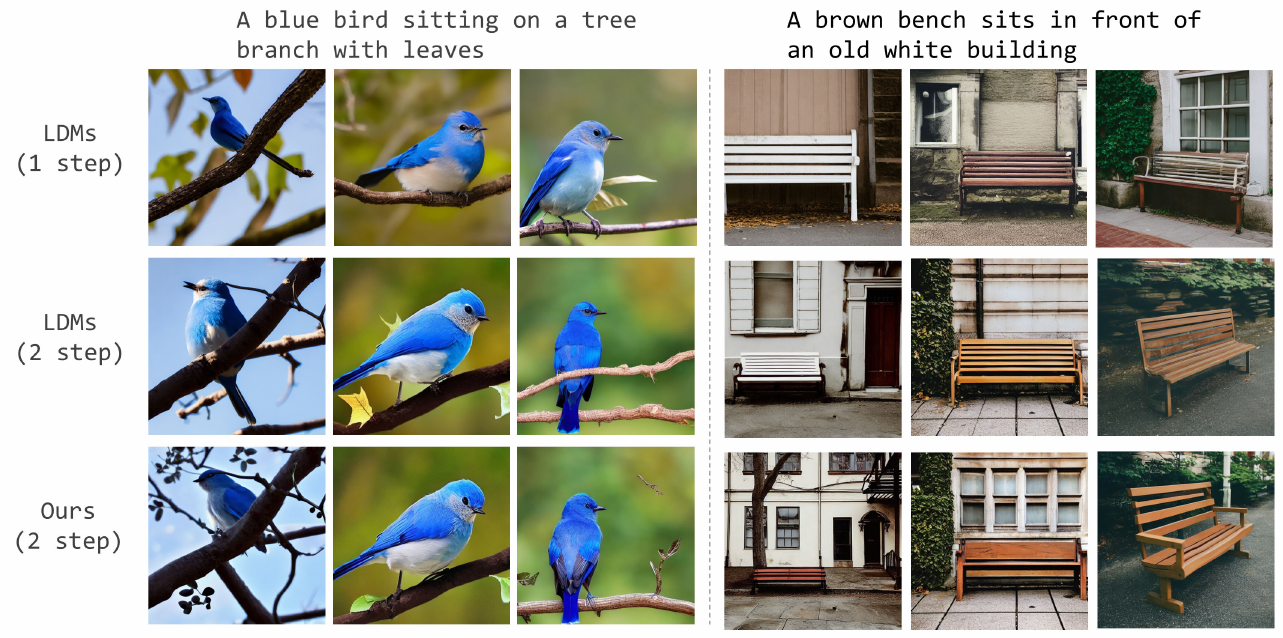}
\caption{The comparison of the images generated with different procedures. '1 step' means the generation is conducted by the 'denoise' process following Eqn.~\ref{equ-stage1}. '2 step' means the generation is conducted by the 'denoise-noise-denoise' process (detail in Section~\ref{exp-relation}). Each column is generated with the same random-sampled noise with the same seed. It can be seen that some results get improved after 2-step denoising, while our method further improves the match with the given texts and the images.}
\label{fig:relation-2step}
\end{figure*}

\subsection{Relational Text}
\label{exp-relation}
In this section, we will show the proposed method on the relational text. Note that the Composable Diffusion~\cite{liu2022compositional} is not suitable for this setting, as the relational words are hard to split. So only LDMs~\cite{rombach2022stable} and Structure Diffusion~\cite{feng2022structure} are incorporated for comparison. The results are shown in Figure~\ref{fig:compare-relation}. Similar to the results of the composable texts, the results of Structure Diffusion~\cite{feng2022structure} still exist the ones that neglect some contents similar to the original LDMs~\cite{rombach2022stable}, while our method successfully synthesizes the missing content, due to the guidance derived from the diffusion models.

Except for generating images from the init noise, here we want to discuss another similar pipeline that can also be used for optimizing the generated images conditioned on the complex texts: When already output $x_0^{fine}$ in Eqn.~\ref{equ-stage1}, the $x_0^{fine}$ are added with the noise again to generate $x_T^{\*}$. Then $x_T^{\*}$ can be used to replace $x_T$ to generate images. For prompt learning, the optimization strategy is the same as before. We denote this 'denoise-noise-denoise' pipeline as '2 step' and the pipeline 'denoise' in Eqn.~\ref{equ-stage1} as '1 step'. The results of '2 step' are shown in Figure~\ref{fig:relation-2step}. It can be seen that, after repeating the denoise-noise procedure, some results get improved, i.e., the second column and the fifth column of LDMs (2 step). After combing with our method, which optimizes the second denoise procedure, the matches between the texts and the images get further improved.

\begin{figure*}[!htbp] 
\centering
\includegraphics[width=1.0\linewidth]{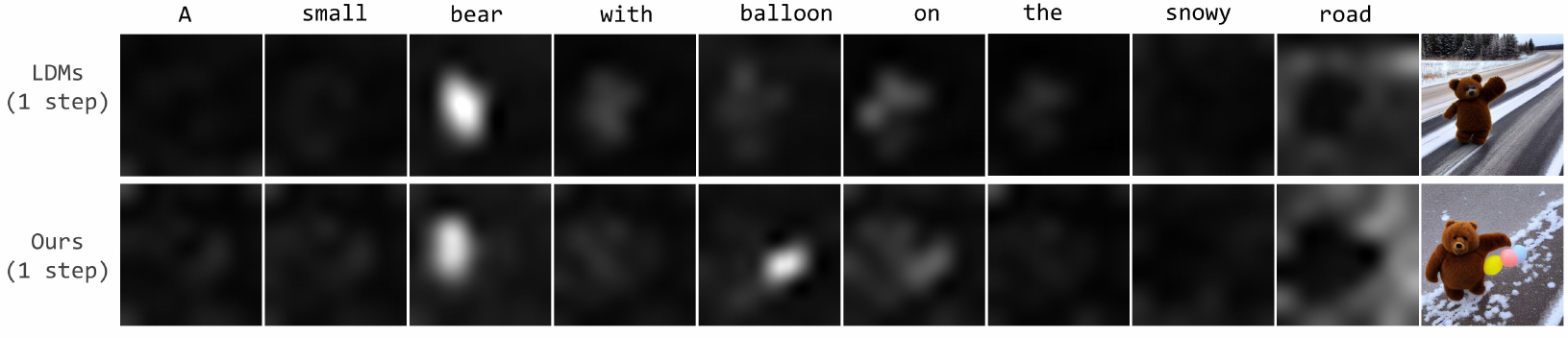}
\caption{The cross-attention maps of the generated images. It can be seen that the neglected objects get enough attention after prompt engineering, leading to accurate text-to-image synthesis. }
\label{fig:heat}
\end{figure*}

\subsection{Intepretable Analysis}
In this section, we conduct a visualizable experiment to show the effect of the learnable prompts. During the denoising process, the input text embeddings interact with the image features through the cross-attention mechanism. Following the methods of ~\cite{hertz2022prompt}, we visualize the cross-attention results as heatmaps to show the attention between the words and the images. The results are shown in Figure~\ref{fig:heat}. It can be seen that the attention to the neglected contents, i.e., 'balloon' and 'in', gets improved after learning proper prompts for them. This is because the texts are encoded using the long-term relationship. Thus, seeking out the proper prompts will improve the modeling of the original texts and provide better conditional guidance for the diffusion model to generate the desired images.

\section{Conclusion}
In this paper, we propose a novel framework to improve text-to-image diffusion synthesis through prompt engineering. Utilizing the quality guidance and the semantic guidance derived from the pre-trained diffusion model, the proposed method successfully seeks out the proper prompts for each textual input. During prompt learning, the parameters of the diffusion model will not be updated, which provides insight into the potential of pre-trained diffusion models to generate better text-conditioned images. Extensive experiments and analyses have validated the effectiveness of our method.

{\small
\bibliographystyle{ieee_fullname}
\bibliography{egbib}
}

\end{document}